# PepHarmony: A Multi-View Contrastive Learning Framework for Integrated Sequence and Structure-Based Peptide Encoding

Ruochi Zhang[1,2,3], Haoran Wu[3], Chang Liu[4], Huaping Li[5], Yuqian Wu[1,8], Kewei Li[1,9], Yifan Wang[3], Yifan Deng[3], Jiahui Chen[3], Fengfeng Zhou[1,9,*], and Xin Gao[6,7,*].

1 Key Laboratory of Symbolic Computation and Knowledge Engineering of Ministry of Education, Jilin University, Changchun, Jilin, China, 130012.

2 School of Artificial Intelligence, Jilin University, Changchun, Jilin, China, 130012.

3 Syneron Technology, Guangzhou, China, 510700.

4 Beijing Life Science Academy, Beijing 102209, China

5 School of Biomedical Sciences, LKS Faculty of Medicine, the University of Hong Kong, Hong Kong SAR, China.

6 Computational Bioscience Research Center, King Abdullah University of Science and Technology (KAUST), Thuwal, Saudi Arabia, 23955.

7 Computer Science Program, Computer, Electrical and Mathematical Sciences and Engeering Division, King Abdullah University of Science and Technology (KAUST), Thuwal, Saudi Arabia, 23955.

8 College of Software, Jilin University, Changchun, Jilin, China, 130012.

9 College of Computer Science and Technology, Jilin University, Changchun, Jilin, China, 130012.

* Corresponding authors.

E-mail addresses: xin.gao@kaust.edu.sa (X. Gao) and FengfengZhou@gmail.com (F. Zhou).

# Abstract

Recent advances in protein language models have catalyzed significant progress in peptide sequence representation. Despite extensive exploration in this field, pre-trained models tailored for peptide-specific needs remain largely unaddressed due to the difficulty in capturing the complex and sometimes unstable structures of peptides. This study introduces a novel multi-view contrastive learning framework PepHarmony for the sequence-based peptide encoding task. PepHarmony innovatively combines both sequence- and structure-level information into a sequence-level encoding module through contrastive learning. We carefully select datasets from the Protein Data Bank (PDB) and AlphaFold database to encompass a broad spectrum of peptide sequences and structures. The experimental data highlights PepHarmony's exceptional capability in capturing the intricate relationship between peptide sequences and structures compared with the baseline and fine-tuned models. The robustness of our model is confirmed through extensive ablation studies, which emphasize the crucial roles of contrastive loss and strategic data sorting in enhancing predictive performance. The proposed PepHarmony framework serves as a notable contribution to peptide representations, and offers valuable insights for future applications in peptide drug discovery and peptide engineering. We have made all the source code utilized in this study publicly accessible via GitHub at https://github.com/zhangruochi/PepHarmony or http://www.healthinformaticslab.org/supp/.

# 1.Introduction

Peptides, as crucial biological entities, embody an array of functionalities in biological processes, ranging from enzymatic activity to therapeutic applications [1, 2]. They are characterized by their shorter amino acid sequences and increased structural flexibility relative to proteins [3]. This distinctiveness positions peptides as key players in numerous biological and therapeutic contexts [4]. However, their unique attributes require tailored computational models for precise representation, an area not yet fully explored by current protein-centric pre-trained models such as ESM(Evolutionary Scale Modeling) [5, 6].

The inherent relationship between a peptide's sequence and its structure is a cornerstone in understanding its function and potential utility [7, 8]. However, representing peptides effectively in computational models, a task fundamental to advancing drug discovery and protein engineering, has been a challenging endeavor. Traditional approaches have either focused on peptide sequences or structures in isolation [9, 10], failing to fully capture the intricate interplay between these two facets.

In recent years, the advent of large-scale protein databases like the Protein Data Bank (PDB) [11, 12] and the transformative AlphaFold DB [13, 14] has provided an unprecedented wealth of data, offering new avenues for peptide research. These developments, alongside advancements in machine learning, particularly in protein pre-trained models [5, 15], have set the stage for innovative approaches in peptide representation.

Our study introduces PepHarmony, a pioneering pretrained model that integrates both peptide sequence and structure information, leveraging the symbiotic relationship between these two aspects. This approach marks a significant departure from traditional models that treat sequence and structure as distinct entities. The crux of PepHarmony lies in its innovative contrastive learning approach, which, during pre-training, fuses structural and sequence information. However, during inference, it exclusively utilizes the sequence encoder. This unique strategy ensures that the extracted features from sequences inherently include structural information, enhancing the model's effectiveness in various peptide-related tasks.

Furthermore, we conduct an exhaustive analysis of peptide sequence and structural data from AlphaFold DB and PDB. This examination allows us to understand better the nuances and intricacies of training data and how they can be effectively harnessed to train our model. We delve into various aspects, such as data distribution, quality, and the impact of different training strategies, providing a comprehensive view of the data landscape.

Our experimental results demonstrate the effectiveness of PepHarmony in various tasks, including peptide-protein interaction prediction and peptide solubility prediction. The model's performance not only attests to its robustness and accuracy but also highlights its potential in facilitating advanced research in protein analysis and drug discovery. In summary, our work not only presents a novel model in PepHarmony but also offers valuable insights into the optimal utilization of peptide data for training.

The three main contributions of our paper are as follows,

1. Introduction of PepHarmony, a pioneering pretrained, pure sequence-based encoder that effectively encapsulates structural data within peptide sequences.

2. Detailed exploration and utilization of peptide sequence and structural data from AlphaFold DB and PDB, including an in-depth analysis of training methodologies and data efficacy.

3. Demonstrated effectiveness of PepHarmony through extensive experimental validation, showcasing superior performance in multiple peptide-related tasks.

## 2. Related work

Understanding the intricate interplay between protein sequences and structures is pivotal in computational biology and bioinformatics [16]. Recent advancements have led to the development of various computational models, each focusing on different aspects of protein analysis. This section reviews the relevant literature, categorized into three main areas: sequence-based protein language models, structure-based protein pretrained models, and models that infuse both structure and sequence information.

### 2.1. Sequence-based protein language models

Protein language models based on sequence data have made significant strides, especially with the adoption of deep learning techniques [17]. Models like BERT [18] and GPT [19], originally designed for natural language processing, have been adapted to interpret the 'language' of proteins. These adaptations, such as ESM, leverage the sequential nature of amino acids to predict protein functionality and interactions. By training on large datasets of known protein sequences, these models have shown remarkable proficiency in capturing the underlying biological properties inherent in sequence data. However, their reliance solely on sequence information limits their ability to fully predict protein behavior, which is also influenced by three-dimensional structural conformations.

### 2.2. Structure-based protein pretrained models

Structure-based models focus on understanding proteins from their three-dimensional conformations. With the advent of more sophisticated structural prediction algorithms and the increasing availability of structural data, these models have gained prominence. Notable among them is AlphaFold [13], which predicts protein structures with remarkable accuracy. Models like GearNet [20] go a step further by incorporating spatial and chemical information into their analyses, providing a more nuanced understanding of protein function and interaction. These models have significantly advanced the field's ability to predict protein functionality based on structure, yet they often overlook the valuable insights that can be gleaned from sequence data.

### 2.3. Structure and sequence infusion models

Recognizing the limitations of focusing on either sequence or structure alone, recent research has seen the emergence of models that integrate both types of information. These multi-view models aim to capitalize on the strengths of both sequence-based and structure-based approaches [20-23]. They involve advanced algorithms that can effectively merge sequence information with structural data, offering a more comprehensive understanding of proteins. The fusion of these two data types allows for a more accurate prediction of protein functions, interactions, and even the discovery of new proteins [24, 25]. This integrated approach represents a significant step forward in the field, providing a holistic view of proteins that is more aligned with their multifaceted nature in biological systems.

However, our work presents distinct approaches and focuses when compared to these existing models. Unlike the model described in [23] which explores the combination of a state-of-the-art Protein Language Model (PLM) and a protein structure encoder, our research concentrates specifically on peptides rather than proteins. [23] demonstrates the efficiency of ESM-GearNet, a model that combines a PLM (ESM-1b) and a protein structure encoder (GearNet) through serial connection and further enhances it with pretraining on massive unlabeled protein structures using contrastive learning. Our study delves into the intrinsic differences between peptide and protein data. We focus on how the unique characteristics of peptides affect the training data and methods used in our models, and how these differences influence the final results. This emphasis on peptides, which are fundamentally different from proteins in terms of size, complexity, and biological roles, requires a tailored approach that differs significantly from the protein-centric models previously developed.

In summary, the field of computational protein analysis is rapidly evolving, with significant contributions from sequence-based, structure-based, and integrated models. Each approach offers unique insights into protein characterization, and the ongoing development of hybrid models promises to unlock even more comprehensive and accurate methods for protein analysis in computational biology.

## 3. Methodology

In the following, we first present an overview of our model, and then individually introduce the models used for sequence representation learning and structure representation learning. Finally, we introduce two learning tasks between sequence and structure.

### 3.1. Overview of multi-view pretrained model

To enhance sequence representation with structural information, our model adopts a multi-view contrastive learning approach. It treats sequence and structure as two complementary views for each peptide (see Figure 1) and performs self-supervised learning (SSL) [26] between these views. In our investigation of sequence encoding from other models, we find that our sequence representation integrates more structural information.

Our model utilizes two pretrained models, ESM and GearNet, for encoding peptide sequences and structures respectively. ESM is a transformer-based sequence encoder and pre-training model, processing peptide sequences to output residue representations. GearNet, used for structure encoding, learns representations that encode spatial and chemical information of peptide structure. The subsequent sections will delve into detailed introductions of these two pretrained models.

At the training stage, our model conducts self-supervised learning. Following the framework in [27], we employ two losses: a contrastive one and a generative one. These losses focus on different representation learning aspects. The contrastive loss concentrates on mapping relationships between sequence and structure representations of peptides. In contrast, the generative loss emphasizes the feature representation of both sequence and structure, particularly the reconstruction of key data features. From a distribution learning perspective, contrastive SSL and generative SSL learn data distribution in a local and global manner, respectively. Contrastive SSL learns the distribution locally by contrasting inter-data pairwise distances, while generative SSL directly learns the global data density function.

## 3.2. Sequence model

ESM, a notable protein language model (PLM), undergoes pre-training using masked language modeling (MLM) loss[18]. This process involves predicting the type of a hidden residue based on its surrounding context. By leveraging a vast amount of unlabeled data, ESM has attained top-tier results in diverse protein understanding tasks. Peptides, essentially shorter sequences of proteins, naturally lend themselves to feature extraction using PLMs like ESM. Despite the absence of a large-scale peptide-specific language model, ESM serves as an effective sequence model encoder. We also utilize the ESM-trained parameters on proteins to initialize the sequence encoder, specifically employing the $ESM_{t12}$ version.

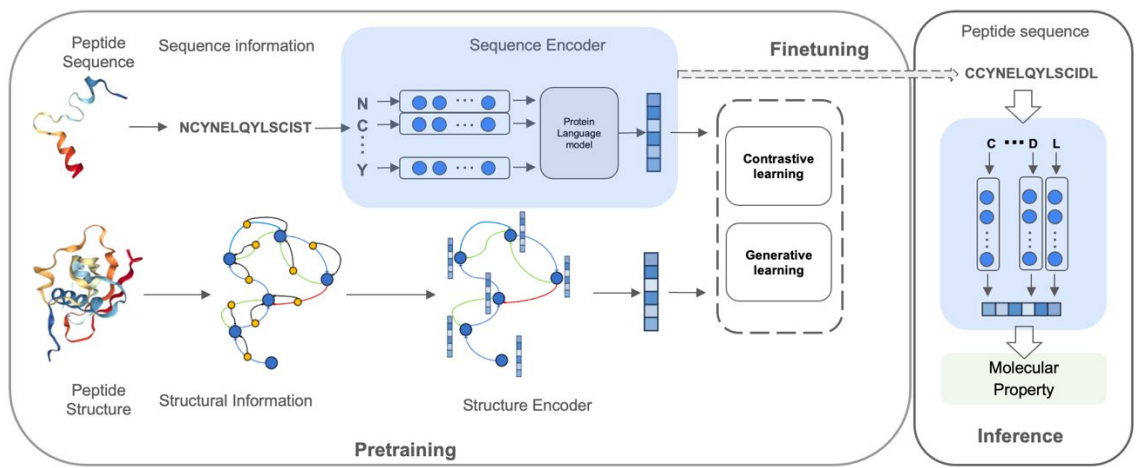

**Figure 1:** Overall architecture of our model. Two encoders, which are sequence encoder and structural encoder reprectively, will be trained together by contrastive or generative learning. After that, in the downstream tasks, we will just use the sequence coder to extract peptide representation.

### 3.3. Structure model

Structures are a direct determinant of peptide functions. We employ a simple yet effective structure-based encoder called GeomEtry-Aware Relational Graph Neural Network (GearNet), which encodes spatial information by adding different types of sequential or structural edges and then performs relational message passing on protein residue graphs. Utilized on our peptide data, GearNet constructs a residue graph for encoding the structure information. For structure model initialization, we employ two versions of pre-trained GearNet, and their performance is extensively compared in our ablation study section. The first version, GearNet$_{cons}$, described in [20], optimizes a contrastive loss function following SimCLR [28], aiming to maximize mutual information between biologically correlated views. It constructs views reflecting protein substructures via two sampling schemes, one being subsequence cropping, which samples residues from a protein graph. The second version, GearNet$_{diff}$, introduced in [29], is inspired by denoising diffusion models' success in generative tasks. This model uses the DiffPreT [29] approach for sequence-structure joint diffusion modeling, guiding the encoder to recover native protein sequences and structures from perturbed ones along a joint diffusion trajectory. It considers protein conformational variations and includes the Siamese Diffusion Trajectory Prediction (SiamDiff) method to capture correlations between different protein conformers.

**Graph construction** Given peptide structures, the representations encoding their spatial and chemical information should be invariant under translations, rotations and reflections in 3D space. To achieve this requirement, GearNet constructs peptide graph based on spatial features invariant under these transformations.

GearNet represents the structure of a peptide as a residue-level relational graph $G = (V, E, R)$, where $V$ and $E$ denotes the set of nodes and edges respectively, and $R$ is the set of edge types. We use $(i, j, r)$ to denote the edge from node $i$ to node $j$ with type $r$. We use $n$ and $m$ to denote the number of nodes and edges, respectively. In this work, each node in the protein graph represents the alpha carbon of a residue with the 3D coordinates of all nodes $x \in \mathbb{R}^{n \times 3}$. We use $f_i$ and $f_{(i,j,r)}$ to denote the feature for node $i$ and edge $(i, j, r)$, respectively, in which reside types, sequential and spatial distances are considered.

Then, we add three different types of directed edges into our graphs: sequential edges, radius edges and K-nearest neighbor edges. Among these, sequential edges will be further divided into 5 types of edges based on the relative sequential distance $d \in \{-2, -1, 0, 1, 2\}$ between two end nodes, where we add sequential edges only between the nodes within the sequential distance of 2. These edge types reflect different geometric properties, which all together yield a comprehensive featurization of proteins.

**Neural message passing architecture** Upon the protein graphs defined above, GearNet utilizes a graph neural network(GNN) [30, 31] to derive per-residue and whole-peptide representations. To balance model capacity and memory cost, GearNet use a relational graph convolutional neural network [32] to learn graph representations, where a convolutional kernel matrix is shared within each edge type and there are $|R|$ different kernel matrices in total. Formally, the relational graph convolutional layer is defined as

$$h_i^{(0)} = f_i, \quad h_i^{(l)} = h_i^{(l-1)} + u_i^{(l)}$$

$$u_i^{(l)} = \sigma\left(BN\left(\sum_{r \in R} W_r \sum_{j \in N_r(i)} h_j^{(l-1)}\right)\right)$$

Specifically, we use node features $f_i$ as initial representations. Then, given the node representation $h_i^{(0)}$ for node i at the l-th layer, we compute updated node representation $h_i^{(l)}$ by aggregating features from neighboring nodes $N_r(i)$, Where $N_r(i) = \{j \in V \mid (j, i, r) \in E\}$ denotes the neighborhood of node i with the edge type r, and $N_r$ denotes the learnable convolutional kernel matrix for edge type r. Here BN denotes a batch normalization layer and we use a ReLU function as the activation $\sigma(\cdot)$. Finally, we update $h_i^{(l)}$ with $u_i^{(l)}$ and add a residual connection from the last layer.

## 3.4. Learning tasks between sequence and structure

In this subsection, we introduce two learning tasks between sequence and structure, contrastive learning and generative learning.

**Contrastive learning** Following the framework of contrastive SSL[28, 33], we first define positive and negative pairs of views from an inter-data level, and then align the positive pairs and contrast the negative pairs simultaneously. For each peptide, we first extract representations from sequence view and structure views, i.e., $h_x$ and $h_y$. Then we create positive and negative pairs for contrastive learning: the sequence-structure pairs (x, y) for the same peptide are treated as positive, and negative otherwise. Finally, we align the positive pairs and contrast the negative ones. The pipeline is shown in Figure 1. In the following, we use InfoNCE objective functions [34] on contrastive graph SSL.

$$L_{InfoNCE} = -\frac{1}{2} E_{p(x,y)} \left[ \log \frac{\exp(f_x(x,y))}{\exp(f_x(x,y)) + \sum_j \exp(f_x(x^j,y))} + \log \frac{\exp(f_y(y,x))}{\exp(f_y(y,x)) + \sum_j \exp(f_y(y^j,x))} \right]$$

where $x^j, y^j$ are randomly sampled 2D and 3D views regarding to the anchored pair (x, y). $f_x(x, y)$ and $f_y(y, x)$ are scoring functions for the two corresponding views, with flexible formulations. Here we adopt $f_x(x, y) = f_y(y, x) = <h_x, h_y>$.

**Generative learning** Generative SSL is another classic track for unsupervised pre-training[35, 36]. It aims at learning an effective representation by self-reconstructing each data point. Specifically to drug discovery, every peptide has a sequence and a structure, and our goal is to learn a robust sequence representation that can, to the most extent, recover its structure counterparts. By doing so, generative SSL can enhance sequence representation to encode the most crucial geometry/topology information, which can improve downstream performance.

GraphMVP [27] proposes a light VAE [37]-like generative SSL, equipped with a crafty surrogate loss. They propose a novel surrogate loss by switching the reconstruction from data space to representation space, called variational representation reconstruction (VRR):

$$L_G = L_{VRR} = \frac{1}{2} \left[ E_{q(z_x|x)} \left[ \|q_x(z_x) - SG(h_y)\|^2 \right] + E_{q(z_y|y)} \left[ \|q_y(z_y) - SG(h_x)\|^2 \right] \right]$$
$$+ \frac{\beta}{2} \cdot \left[ KL(q(z_x|x) \| p(z_x)) + KL(q(z_y|y) \| p(z_y)) \right].$$

In this formula, SG represents the stop-gradient operation, which is based on the assumption that $h_y$ is a fixed, learned representation function. The use of SG is a common practice in SSL literature, primarily to prevent model collapse, as noted in [38].

## 4. Experiments

### 4.1. Dataset Acquisition

**PDB dataset**: The Protein Data Bank (PDB) is a globally accessible public database that offers detailed information on the three-dimensional structures of various biological macromolecules, such as proteins, nucleic acids, and complex assemblies. These structures are made available without charge to the international research community, supporting a broad spectrum of scientific endeavors including drug discovery, protein engineering, and fundamental studies of biological molecules' structure and functionality. In our specialized utilization of the PDB, we extract all protein chains from the PDB database and assessed them based on the sequence length of their amino acids, identifying chains as peptides if their length was under 50. Through this meticulous process, we successfully isolated and cleansed data for 17,300 peptides, including both their amino acid sequences and corresponding structural information.

**AlphaFold dataset:** AlphaFold DB, capitalizing on the advanced AlphaFold model, provides open access to more than 200 million protein structure predictions, markedly accelerating scientific research. It offers an extensive selection of predicted protein structures, easily accessible to the scientific community. Drawing from AlphaFold DB, we apply a methodology akin to our PDB data cleansing strategy, successfully extracting structural information for 3.5 million peptide sequences.

The AlphaFold DB and PDB datasets both contain structural data for peptides and sequences. However, they exhibit distinct data distributions. As shown in Figure 2(a), the peptide length distribution in the PDB dataset is significantly shorter compared to that in the AlphaFold DB. Acknowledging that the structures predicted by AlphaFold2 are theoretical rather than actual, we theorized that the precision of these predictions— essentially the data's quality—would profoundly influence the training of models. AlphaFold computes a per-residue confidence score from 0 to 100, known as pLDDT (predicted Local Distance Difference Test), which mirrors the model's confidence in its predictions. In our approach, we averaged these pLDDT scores for each peptide, yielding a single confidence score indicative of the overall reliability of each peptide's predicted structure. As depicted in Figure 2(b), we note a trend: shorter sequences typically have higher pLDDT score distributions. This led us to an insightful strategy: by selecting training data with higher pLDDT scores, we could align the length distribution of this subset more closely with that observed in the PDB datasets.

Consequently, we categorized our training dataset into distinct quality datasets based on pLDDT scores for more effective experimental validation:

**af90:** About 500,000 data samples were selected, each with a confidence score above 90.

**af80:** Approximately 1.18 million data samples were chosen, all with confidence scores above 80.

**af50w:** A varied set of 500,000 data samples was randomly sampled from the entire dataset, irrespective of confidence scores.

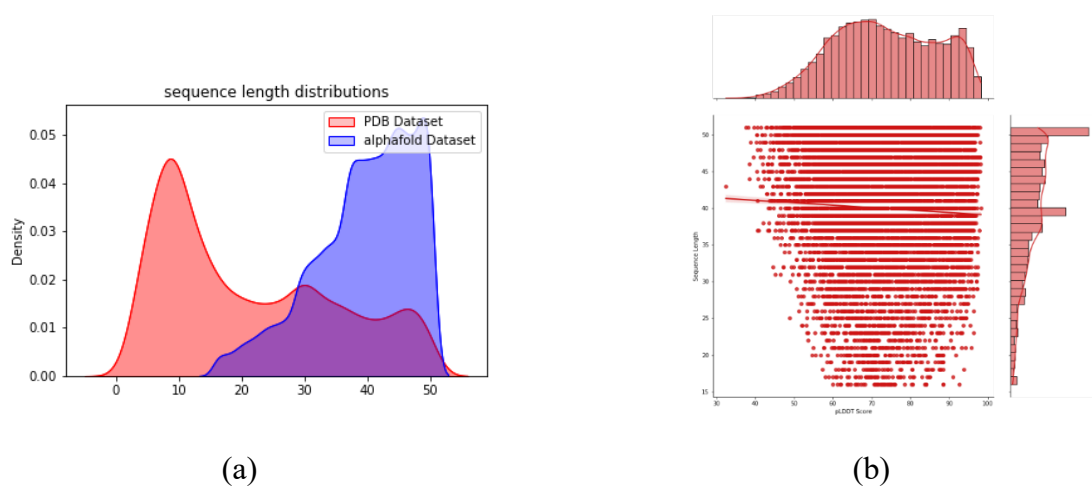

(a) (b)

**Figure 2:** Statistaics for PDB and AlphaFold DB datasets. (a) Illustrates the differences in peptide length distribution between the PDB Dataset and the AlphaFold Dataset. Overall, the peptide lengths in the PDB Dataset are shorter than those in the AlphaFold DB. (b) Shows the relationship between the averaged pLDDT scores and the peptide lengths in the AlphaFold Dataset. As can be seen from the figure, shorter sequences typically exhibit higher distributions of pLDDT scores.

This by pLDDT scores enable a more nuanced and quality-focused selection of data for training and research purposes. For our model training, we only utilize data from the AlphaFold DB, while for testing, we employed the PDB datasets. Since the PDB datasets are derived from actual experimental results, using them for self-contaction evaluation ensures a more equitable assessment. We also explored the approach of mixing data from both the AlphaFold DB and PDB datasets for combined training and testing purposes.

**CPP dataset** We first evaluate our models on the dataset of CPP [39-42]. Our CPP dataset is the largest known database of cell-penetrating peptides, containing 1162 positive and negative samples each. We sorted out 22 relevant cell-penetrating peptide databases by compiling literature on existing cell-penetrating peptide prediction models. To ensure the reliability of the sources, we traced each database and listed the relationships between the databases. Besides, we also checked whether there were conflicts and duplicates in different databases, and the anomalous data were checked artificially again.

**Solubility dataset** We utilize PROSO-II dataset [43] for our peptide solubility prediction task. Thanks to the restrictive data selection from the pepcDB and PDB databases, PROSO-II dataset is one of the largest available databases used for solubility-model building and evaluation. A large portion of the data comes from pepcDB database, in which each protein is associated with multiple amino acid sequences corresponding to different constructs. For each construct, its status history is recorded, including the one of solubility. Hence, all constructs that achieved the soluble status are considered as positive samples. Besides, additional soluble data originates also from PDB entries of heterologous complexes, the proteins with the annotation of 'Expression Organism: ESCHERICHIA COLI' are considered positive. Adapting PROSO-II dataset to our needs, we have constructed a subset (about 4 thousand pieces of data) which contains only sequences less than 50 in length.

**Affinity dataset** We construct our benchmark data set from two sources: first is the protein–peptide complex structures from PDB database and second is the drug-target pairs from DrugBank [44-48]. In total, we obtained 7417 positive interacting pairs covering 3412 protein sequences and 5399 peptide sequences. Among them, 6581 pairs from PDB have residue-level binding labels in peptide sequences. We then constructed a negative dataset by randomly shuffling those non-interacting pairs of proteins and peptides. More specifically, for each positive interaction, five negatives were generated by randomly sampling from all the shuffled pairs of non-interacting proteins and peptides. Overall, we obtained 44,502 peptide–protein pairs in our benchmark.

## 4.2. Experimental Setup

**Training details** In the training phase of our model, we explore the effects of various components on the outcomes. We utilized the pre-trained ESM model for sequence embedding and two versions of the pre-trained GearNet model for structure embedding. The emphasis was on using contrastive learning loss for its intuitive alignment of peptide sequences with their structures. Additionally, we assessed the influence of generative learning loss on training outcomes. During pretraining, we noted a significant drop in validation loss and high accuracy with limited data, raising concerns about data leakage. This was attributed to GearNet potentially leaking sequence information into structural representations and the model's ability to match sequence and structure pairs based on length. To mitigate these, we masked amino acid information in GearNet and altered batch selection, reducing accuracy to about 0.75.

**Downstream tasks** This work aims to incorporate structural information into the sequence representation of peptides through contrastive learning. To achieve this, we extract the sequence encoding model and design the following four downstream tasks to validate the learned sequence and structure information of the model.

- **CPP:** penetrating peptide prediction (binary classification).

- **Solubility:** peptide solubility prediction (binary classification).

- **Affinity:** peptide-protein affinity prediction (regression).

- **Self-Contaction:** self-contact map prediction (binary matrix prediction). This task focuses on peptide structure, predicting the connectivity between amino acids.

**Focus on sequence encoder in downstream tasks** Although the contrastive learning framework theoretically allows for learning two independent encoders: the sequence encoder and the structure encoder, our focus in a downstream task is solely on the performance of the sequence encoder. Specifically, we use amino acid sequences as input to extract representations and predict the properties of peptides. The reasons for this focus are threefold:

1. Our aim is to effectively integrate structural information into the sequence model through the use of contrastive learning or a generative learning framework.
2. The sequence encoder has a high throughput [49], making it particularly useful in new drug development scenarios for applications like virtual screening [50]. It has a wide range of potential applications.
3. In most downstream scenarios, especially in the early stages of new drug development, when predicting the properties of a peptide, we usually only have sequence information and not structural information [51]. Therefore, the structure encoder is unable to extract features due to the absence of structural data.

## 4.3. Dataset efficacy in predictive modeling

The results presented in Figure 3 indicate a comparative analysis of validation performance using three distinct datasets derived from AlphaFold: af50w, af80, and af90, each corresponding to different confidence levels. The datasets are evaluated on two downstream tasks: CPP (protein-protein interaction prediction) and Affinity(protein-ligand affinity prediction).

For the CPP task, the validation loss and AUC-ROC metrics are considered. As depicted in Figure 3(a) and (b), the af90 dataset demonstrates a marginally lower validation loss and a higher AUC-ROC compared to af50w andaf80. These differences suggest that the af90 dataset enables the model to predict protein-protein interactions with greater accuracy and reliability.

In the context of the Affinity task, Figure 3(c) and Figure 3(d) display the validation loss and Pearson Correlation coefficient, respectively. Consistent with the CPP task results, the af90 dataset exhibits the lowest validation loss and the highest Pearson Correlation coefficient, reinforcing the assertion that higher confidence in the pre-training dataset correlates with superior model performance on downstream tasks.

The experimental outcomes robustly support the conclusion that the af90 dataset, characterized by the highest confidence level, yields the best performance across both downstream tasks. This affirms the hypothesis that the quality of the pre-training dataset is crucial for enhancing model efficacy. The slight but consistent superiority of af90 over the other datasets across all metrics provides empirical evidence that higher confidence scores are significantly beneficial for the model's predictive capabilities.

Our subsequent investigations focused on leveraging the high-quality, experimentally measured data available in the PDB dataset. Initially, we hypothesized that a combination of PDB's precision with the extensive volume of af90 might yield optimal results. To this end, we implemented two data integration strategies: a direct mix strategy and a sequential finetune strategy.

The mix strategy involved an indiscriminate combination of data from both sources for the training process. In contrast, the finetune strategy entailed initial model pre-training on the af90 dataset followed by refinement using the PDB dataset. These strategies were predicated on the notion that the PDB's data, despite its smaller size, would infuse the model with experimentally validated nuances, thereby enhancing the overall model performance.

Contrary to our expectations, the results depicted in Figure 4 demonstrate that neither data mixing strategy provided any significant advantage over the exclusive use of af90. The af90 dataset emerged as the clear frontrunner, outperforming the mixed strategies on both CPP and Affinity tasks. This suggests that the quantity of data in af90, which surpasses that of the PDB by an order of magnitude, plays a more critical role in model training than the exclusively high-quality yet limited PDB data. The sheer volume of af90 appears to facilitate a more comprehensive learning of peptide representations, despite its marginally inferior data quality compared to the PDB.

Moreover, the underperformance of the mix and finetune strategies may be attributed to the heterogeneous nature of the datasets. The PDB and AlphaFold databases exhibit significant differences in their data distributions, a factor that potentially undermines the efficacy of mixed training sets. These distributional disparities have been extensively discussed in earlier sections of this study.

In light of these findings, we have elected to employ the af90 dataset as the training foundation for PepHarmony, our model of choice for subsequent experiments. The ensuing chapters will delve into the implications of this decision and the resulting performance enhancements in peptide representation learning.

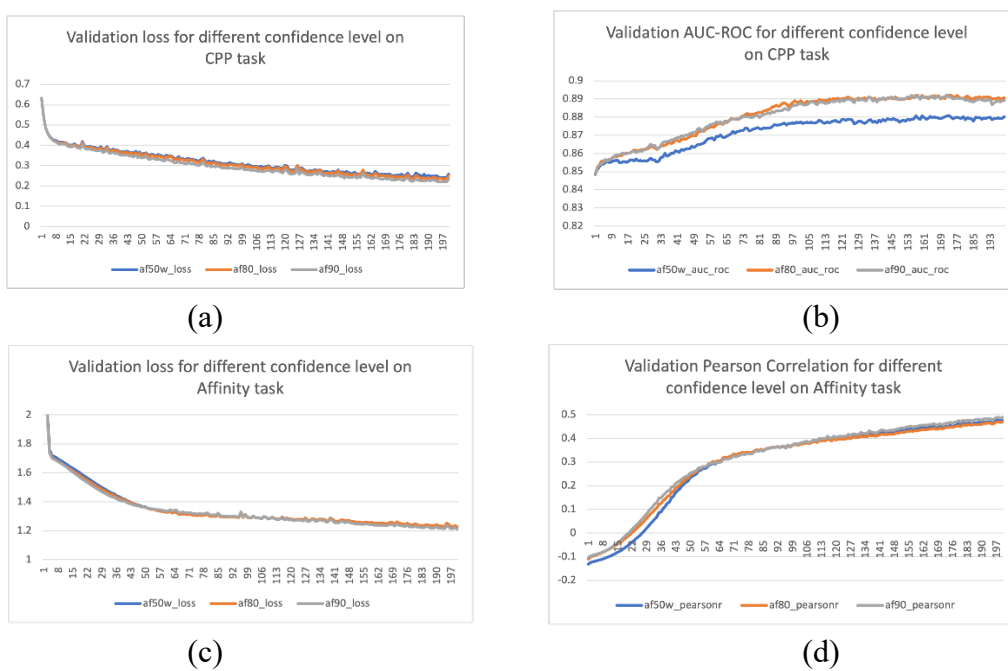

**Figure 3:** The impact of AlphaFold DB data with different confidence levels on downstream tasks.(a) The model trained with af90 achieved the lowest validation loss in the CPP task. (b) Both af90 and af80 models obtained similar validation AUC-ROC scores in the CPP task, significantly outperforming af50. (c) The model trained with af90 also achieved the lowest validation loss in the affinity task. (d) Moreover, the af90-trained model scored the best Pearson correlation in the affinity task. Combining all the results, it indicates that high-quality data positively impacts the final model's results.

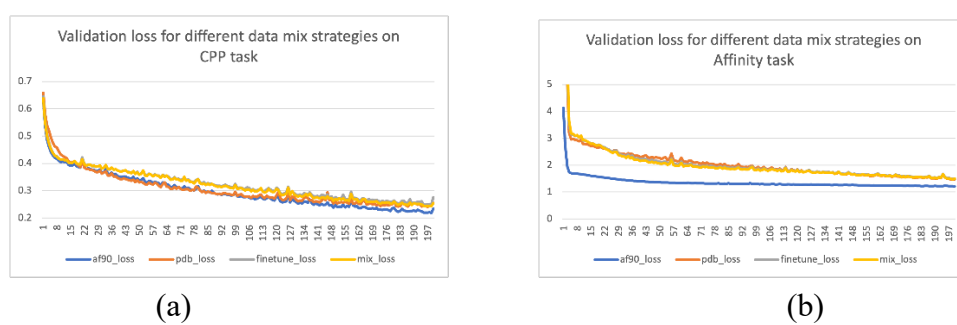

**Figure 4:** Performance of different datasets mix methods on two downstream tasks. (a) In the CPP task, the model trained on the af90 dataset achieves the best validation loss, outperforming both data mix strategies and the PDB dataset. (b) In the Affinity task, the model trained on the af90 dataset not only achieves the best validation loss but also exhibits a very rapid convergence rate.

## 4.4. Overall performance and comparison

In this section, we present a quantitative analysis of our novel multi-view contrastive learning model, PepHarmony, against several baseline models. The evaluation metrics include accuracy (ACC), F1 score, receiver operating characteristic area under the curve (ROC-AUC), root mean square error (RMSE), and correlation coefficients (Pearson and Spearman), across four different tasks: CPP, Solubility, Affinity, and Self-contact prediction.

Our PepHarmony$_{cl}$ model, utilizing only contrastive loss, demonstrated superior performance in most tasks, outstripping the baseline and fine-tuned models. As can be seen in Table 1, PepHarmony$_{cl}$ showed remarkable improvements, particularly in the CPP task, with an ACC of 0.79 and an F1 score of 0.766, which are the highest among the compared models. Furthermore, it achieved the highest ROC-AUC scores across all tasks, highlighting its robust predictive capabilities.

The comparison between the original ESM2 model and its fine-tuned counterpart (ESM$_{finetune}$) illustrates the benefits of fine-tuning with peptide sequence data, as evidenced by the uplift in performance across all tasks. The ESM$_{finetune}$ model, for instance, shows an increase in ACC from 0.723 to 0.77 in the CPP task. Similarly, the GearNet$_{finetune}$ model demonstrates the advantages of fine-tuning with peptide structural data, showing a substantial increase in performance metrics compared to the original GearNet model.

Among the variations of PepHarmony, the VAE model, which incorporates both contrastive and generative loss, also shows promising results, often outperforming the baseline models and coming second only to the PepHarmony$_{cl}$ model. It suggests that the integration of generative loss with contrastive loss can enhance learning peptide representations but does not necessarily translate to surpassing the contrastive loss model in all tasks.

In conclusion, our PepHarmony$_{cl}$ model emerges as the leading model for peptide representation learning, excelling in capturing the complex interplay between peptide sequences and structures. This superiority in learning efficacy makes it the best pre-trained peptide model to date, as it effectively utilizes high-confidence data and advanced dataset mixing methods.

Table 1. Results for four downstream tasks.

|              | ACC   | F1    | ROC-  | ACC   | F1    | ROC-  |
|--------------|-------|-------|-------|-------|-------|-------|
| ESM2         | 0.723 | 0.756 | 0.864 | 0.586 | 0.618 | 0.610 |
| ESM_finetune | 0.770 | 0.755 | 0.839 | 0.645 | 0.663 | 0.666 |
| GearNet      | 0.718 | 0.734 | 0.822 | 0.584 | 0.616 | 0.608 |
| GearNet_finetune | 0.764 | 0.753 | 0.842 | 0.612 | 0.706 | 0.674 |
| PepHarmony_cl | **0.790** | **0.766** | **0.874** | **0.645** | **0.724** | **0.692** |
| PepHarmony_vae | 0.777 | 0.764 | 0.871 | 0.638 | 0.721 | 0.682 |
|  | \multicolumn{3}{c}{Affinity} | \multicolumn{3}{c}{Self-contaction} |
|  | RMSE | Pearson | Spearman | ACC | F1 | ROC- |
| ESM2 | 1.360 | 0.465 | 0.428 | 0.657 | 0.710 | 0.790 |
| ESM_finetune | 1.327 | 0.493 | **0.473** | 0.656 | 0.712 | 0.765 |
| GearNet | 1.343 | 0.452 | 0.426 | 0.637 | 0.712 | 0.723 |
| GearNet_finetune | 1.336 | 0.473 | 0.444 | 0.661 | **0.717** | 0.745 |
| PepHarmony_cl | **1.302** | **0.499** | 0.447 | **0.669** | 0.711 | **0.796** |
| PepHarmony_vae | 1.306 | 0.495 | 0.443 | 0.661 | 0.711 | 0.764 |

## 4.5. Model interpretability

In the pursuit of assessing the efficacy of our novel algorithm PepHarmony in the context of proteomic analysis, we perform comparative experiments to scrutinize the capabilities of the model in fusing structural and sequence information within its representation space. This exploration aims to substantiate the model's proficiency in delineating distinct biological tissue features across varying scales of complexity. To this end, we harness a peptide dataset with less than 50 amino acids in length from the SCOP database, resulting in a dataset of 577 sequences spanning 393 protein families. To streamline our analysis, we filter out families represented by fewer than 10 peptide sequences. Consequently, our final dataset comprises 577 sequences belonging to 27 distinct protein families.

In our methodological approach, we conscientiously addressed the influence of sequence homogeneity by implementing a clustering algorithm specifically designed to group peptides by their sequence similarity. This strategic clustering facilitated the selection of a dataset subset characterized by peptides sharing high sequence resemblance yet exhibiting pronounced structural diversity. The result of this selection process is graphically represented in the accompanying figure, where each color corresponds to a unique protein family and each marker shape represent a cluster, thereby underscoring the structural variance despite the considerable sequence similarity among the peptides.

Our analysis, particularly evident in Figure 5(b), highlights PepHarmony's enhanced capability to discriminate between different protein families. Notably, in the right half of Figure 5(b), PepHarmony distinctly separates peptides belonging to different families, whereas in the corresponding section of Figure 5(a), these peptides are indistinguishably mixed. This observation underscores our model's advanced learning and integration of structural information. Similarly, in the upper half of the diagram, where all data points are circular, indicating high sequence similarity, PepHarmony still successfully distinguishes the blue sequences. This demonstrates that despite high sequence similarity, PepHarmony can effectively differentiate sequences based on their structural characteristics.

Through this comparative visualization, we underscore PepHarmony's superior capability in capturing and integrating the multifaceted nuances of peptide structures, thereby reinforcing its potential as a useful tool in protein sequence analysis.

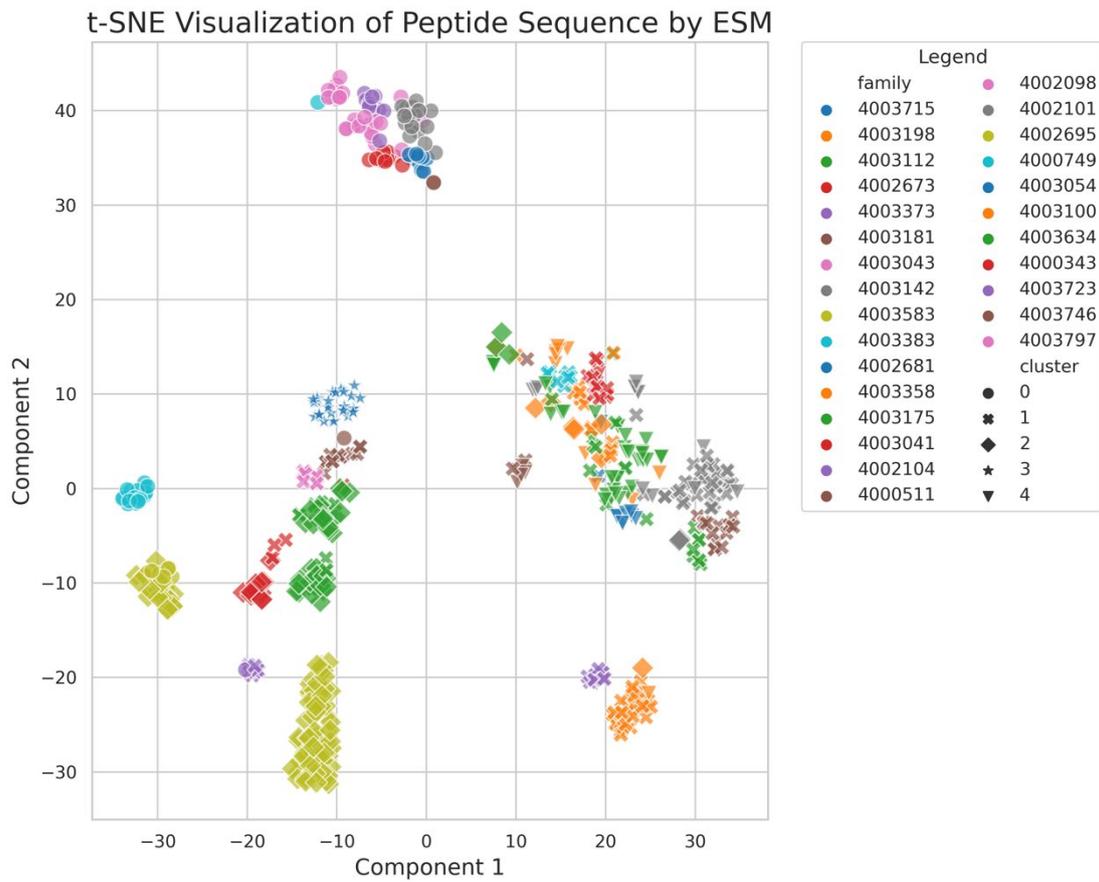

(a)

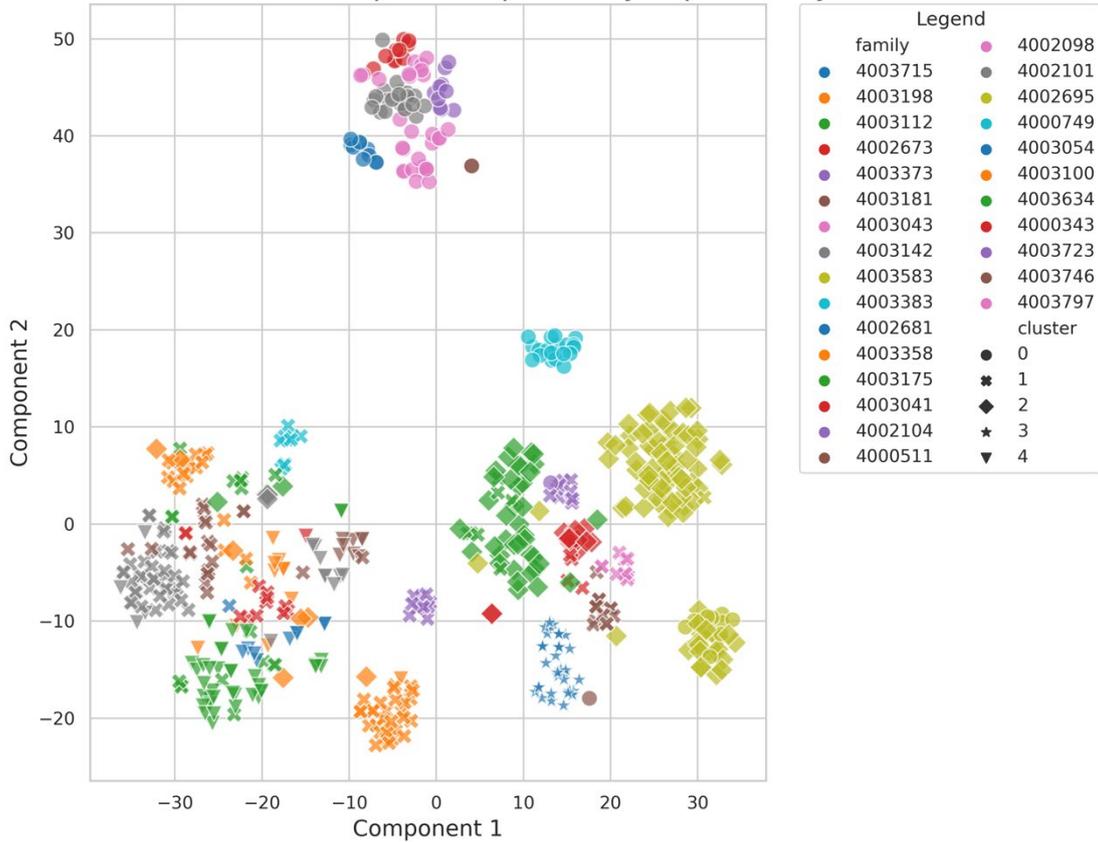

(b)

**Figure 5:** The distribution of peptide representations obtained by different algorithms after dimensionality reduction with t-SNE in two-dimensional space. (a) Representation of ESM in two-dimensional space; (b) Representation of PepHarmony in two-dimensional space.

## 4.6. Ablation experiments

In the ablation study presented in Table 1, we differentiate between several key components of our training methodology. The 'Contrastive' column denotes the application of contrastive learning tasks, which are designed to enhance the discriminative power of the model by contrasting positive and negative examples. The 'Generative' column indicates the use of generative learning tasks, which aim to generate new data instances that are consistent with the training data distribution. The 'Contrastive pre-trained GearNet' and 'Diffusion pre-trained GearNet' columns identify the initialization strategies for our model. 'Contrastive pretrained GearNet' refers to the model being initialized with parameters obtained from a GearNet model pre-trained with contrastive learning tasks. Similarly, 'Diffusion pre-trained GearNet' pertains to the model being initialized with parameters from a GearNet model that has undergone pretraining with a diffusion-based learning process. Lastly, 'Data sort' specifies our approach to batch construction. By sorting peptide sequences by length before batching, we aim to minimize padding and improve computational efficiency. Additionally, this approach is intended to prevent data leakage by ensuring that similar sequences are processed together, thereby preventing the model from inadvertently learning from the batch structure.

Table 3 presents our comprehensive findings across four key downstream tasks. These results allow us to draw several critical conclusions:

**Effectiveness of contrastive loss and Data sorting:** The $CL_{con}$ sort configuration, which combines contrastive loss with data sorting, shows superior performance in the CPP task, achieving the highest ACC, F1, and ROC-AUC scores (0.79, 0.766, and 0.874, respectively). This configuration also maintains consistent performance in the Solubility task. The improvement in metrics, especially in the CPP task, underscores the efficacy of integrating contrastive loss with a data sorting strategy.

**Comparative analysis of pre-trained models:** When comparing the pre-trained models, the Contrastive pretrained GearNet ($CL_{con}$ and $CL_{con\_sort}$) consistently outperforms the Diffusion pre-trained GearNet ($CL_{diff\_sort}$) across most metrics in different tasks. For instance, in the Affinity task, $CL_{con\_sort}$ achieves a lower RMSE (1.302) and higher Pearson (0.499) and Spearman (0.447) scores compared to $CL_{diff\_sort}$. This indicates a clear advantage of using a contrastive learning-based pre-training over diffusion-based pre-training for GearNet.

**Limited impact of VAE loss:** The inclusion of VAE loss ($CL_{vae\_con\_sort}$) results in marginal improvements in some tasks, such as a slightly higher F1 score in Solubility (0.721). However, its overall impact across tasks is limited compared to the other configurations, particularly in the Affinity and Self-contaction tasks, where its performance is comparable or slightly lower than the other models.

In summary, our ablation experiments clearly demonstrate the superiority of using contrastive loss in combination with a data sorting strategy for training the model. Additionally, they reveal that a Contrastive learning pre-trained GearNet is more effective than a Diffusion pre-trained GearNet for the tasks examined. The limited contribution of VAE loss to overall model performance suggests its lesser relevance in the context of our specific application.

Table 2. Ablation on the pre-train learning tasks and pretraining parameters.

|  | Contrastive | Generative | Contrastive pre-trained GearNet | Diffusion pre-trained GearNet | Data sort |
|---|---|---|---|---|---|
| CL_con | ✓ |  | ✓ |  |  |
| CL_con_sort | ✓ |  | ✓ |  | ✓ |
| CL_diff_sort | ✓ |  |  | ✓ | ✓ |

| | | | | | | ✓ |
|---|---|---|---|---|---|---|
| CL_vae_con_sort | ✓ | ✓ | ✓ | | | ✓ |

Table 3. Results for Ablation experiments on four downstream tasks.

| | CPP | | | Solubility | | |
|---|---|---|---|---|---|---|
| | ACC | F1 | ROC-AUC | ACC | F1 | ROC-AUC |
| CL_con | 0.773 | 0.762 | 0.843 | 0.645 | 0.693 | 0.691 |
| CL_con_sort | **0.790** | **0.766** | **0.874** | **0.645** | 0.672 | **0.692** |
| CL_diff_sort | 0.764 | 0.753 | 0.855 | 0.625 | **0.693** | 0.644 |
| CL_vae_con_sort | 0.777 | 0.764 | 0.871 | 0.638 | 0.721 | 0.682 |
| | Affinity | | | Self-contaction | | |
| | RMSE | Pearson | Spearman | ACC | F1 | ROC-AUC |
| CL_con | 1.347 | 0.445 | 0.409 | 0.657 | **0.713** | 0.788 |
| CL_con_sort | 1.302 | **0.499** | **0.447** | **0.669** | 0.711 | **0.796** |
| CL_diff_sort | **1.299** | 0.778 | 0.449 | 0.655 | 0.705 | 0.769 |
| CL_vae_con_sort | 1.306 | 0.495 | 0.443 | 0.661 | 0.711 | 0.764 |

## 5. Conclusions and future work

In this study, we introduce PepHarmony, a pioneering model in the field of peptide representation learning, notable for its unique identity as a pure sequence-based encoder. Utilizing a novel multi-view contrastive learning approach, PepHarmony successfully integrates sequence and structural information in its training phase. However, it's crucial to note that during inference, the model exclusively leverages its sequence encoder. This design choice is not merely a technical detail but a strategic innovation, allowing the model to extract features that inherently encapsulate structural information, leading to exceptional results in various analytical tasks.

Through extensive experimentation, we have demonstrated that the quality and preparation of training data play a pivotal role in model performance. Our exploration of datasets from AlphaFold DB and PDB revealed key insights into data distribution, quality, and training strategies. The adept handling of these datasets has been instrumental in harnessing their full potential for training PepHarmony. Our ablation studies further delineated the importance of contrastive learning and the integration of different pretraining strategies. The outcomes suggest that while the incorporation of generative loss provides certain benefits, the model predominantly excels with contrastive loss, especially when coupled with data sorting techniques.

As we move forward, several avenues for future work emerge. First, the exploration of domain-specific adaptations of PepHarmony could be pursued to tailor the model for specific applications, such as targeted drug design or protein engineering. Second, integrating emerging datasets and continual advancements in protein structure prediction methods could further enhance the model's robustness and accuracy. Finally, exploring the potential of transfer learning and its implications in rapidly evolving fields like synthetic biology could open up new frontiers for PepHarmony.

In conclusion, PepHarmony represents a significant step forward in peptide representation learning. Its ability to effectively capture the intricate relationship between peptide sequences and structures offers a powerful tool for researchers and practitioners alike. We anticipate that this model will serve as a useful tool for future innovations in the field, contributing to advancements in drug discovery, protein engineering, and beyond.

## Acknowledgements

This publication is based upon work supported by the King Abdullah University of Science and Technology (KAUST) Office of Research Administration (ORA) under Award No URF/1/4352-01-01, FCC/1/1976-44-01, FCC/1/1976-45-01, REI/1/5234-01-01, REI/1/5414-01-01, REI/1/5289-01-01, REI/1/5404-01-01.

This work is also supported by the Senior and Junior Technological Innovation Team (20210509055RQ), the Jilin Provincial Key Laboratory of Big Data Intelligent Computing (20180622002JC), and the Fundamental Research Funds for the Central Universities, JLU.